\documentclass[conference]{IEEEtran}
\usepackage{cite}
\usepackage{amsmath,amssymb,amsfonts}
\usepackage{algorithmic}
\usepackage{graphicx}
\usepackage{textcomp}
\usepackage{xcolor}
\usepackage[hyphens]{url}
\usepackage{hyperref}

\usepackage{microtype}
\usepackage{subfigure}
\usepackage{booktabs}
\usepackage{multirow}
\usepackage{makecell}
\usepackage[most]{tcolorbox}
\usepackage{etoolbox}
\usepackage[utf8]{inputenc}
\usepackage{authblk}

\usepackage{listings}
\usepackage{amsmath}
\usepackage{amssymb}
\usepackage{bbm}
\usepackage{float}
\usepackage[htt]{hyphenat}

\newcommand{\modelname}{KForge}

\def\BibTeX{{\rm B\kern-.05em{\sc i\kern-.025em b}\kern-.08em
    T\kern-.1667em\lower.7ex\hbox{E}\kern-.125emX}}

\usepackage{yfonts}
\title{KForge: LLM-Driven Cross-Platform Kernel Generation for AI Accelerators}

\author{%
\textbf{Taras Sereda}\quad
\textbf{Burak Bartan}\quad
\textbf{Ankita Nayak}\quad
\textbf{Tom St.~John}\\
\textbf{Natalie Serrino}\quad
\textbf{Zain Asgar}\\
Gimlet Labs Inc., San Francisco, California, USA\quad
\\
\footnotesize{\texttt{{\{\href{mailto:taras@gimletlabs.ai}{taras},\,\href{mailto:bbartan@gimletlabs.ai}{bbartan},\,\href{ankitanayak@gimletlabs.ai}{ankitanayak},\,\href{mailto:tstjohn@gimletlabs.ai}{tstjohn},\,\href{mailto:nserrino@gimletlabs.ai}{nserrino},\,\href{mailto:zasgar@gimletlabs.ai}{zasgar}\}@gimletlabs.ai}}}
}
\usepackage{fancyhdr}
\fancypagestyle{firstpage}{
  \fancyhf{} 
  \chead{\small \textit{Accepted to the Machine Learning for Architecture and Systems Workshop (\href{https://sites.google.com/corp/view/mlarchsys}{MLArchSys}), co-located with \href{https://iscaconf.org/isca2026/}{ISCA 2026}.}}
  \cfoot{\thepage} 
}

\begin{document}
\maketitle

\thispagestyle{firstpage} 
\pagestyle{plain}


\begin{abstract}
Production inference increasingly targets a heterogeneous mix of accelerators. Agentic pipelines interleave reasoning, tool calls, and multi-agent coordination, each with distinct compute and memory profiles. For optimal efficiency, each stage should run on the accelerator best suited to it.
This creates a systems challenge: each pipeline now requires high-performance kernels across a growing set of hardware backends and programming models. Writing these kernels by hand is time-consuming, demands deep low-level expertise, and does not scale as kernel complexity grows. Recently, Large Language Models (LLMs) have been leveraged for automatic kernel generation, but challenges in low-level code generation and cross-backend generalization  persist. We present \modelname{}, a cross-platform framework built around an iterative refinement loop driven by two collaborating LLM-based agents: a generation agent that produces and progressively refines kernels using compilation and correctness feedback, and a performance-analysis agent that interprets profiling data, from programmatic APIs to GUI-based tools, and emits recommendations that steer the next round of synthesis. The loop alternates between functional passes, which drive a candidate to correctness, and optimization passes, which close the performance gap to hand-tuned baselines. We evaluate KForge on two backends with very different baseline reference availability. On NVIDIA B200, \modelname{} achieves a 2.12\% improvement in end-to-end throughput compared to TensorRT-LLM on the gpt-oss-20b inference speed benchmark. On Intel Arc B580, \modelname{} generates Triton kernels achieving a 5.13$\times$ geometric mean speedup over the faster of PyTorch eager and \texttt{torch.compile} on 37 GEMM + tail-ops workloads from KernelBench Level~2, primarily via operator fusion and mixed-precision execution.

\end{abstract}

\begin{figure*}[t]
    \centering
    \includegraphics[scale=0.45]{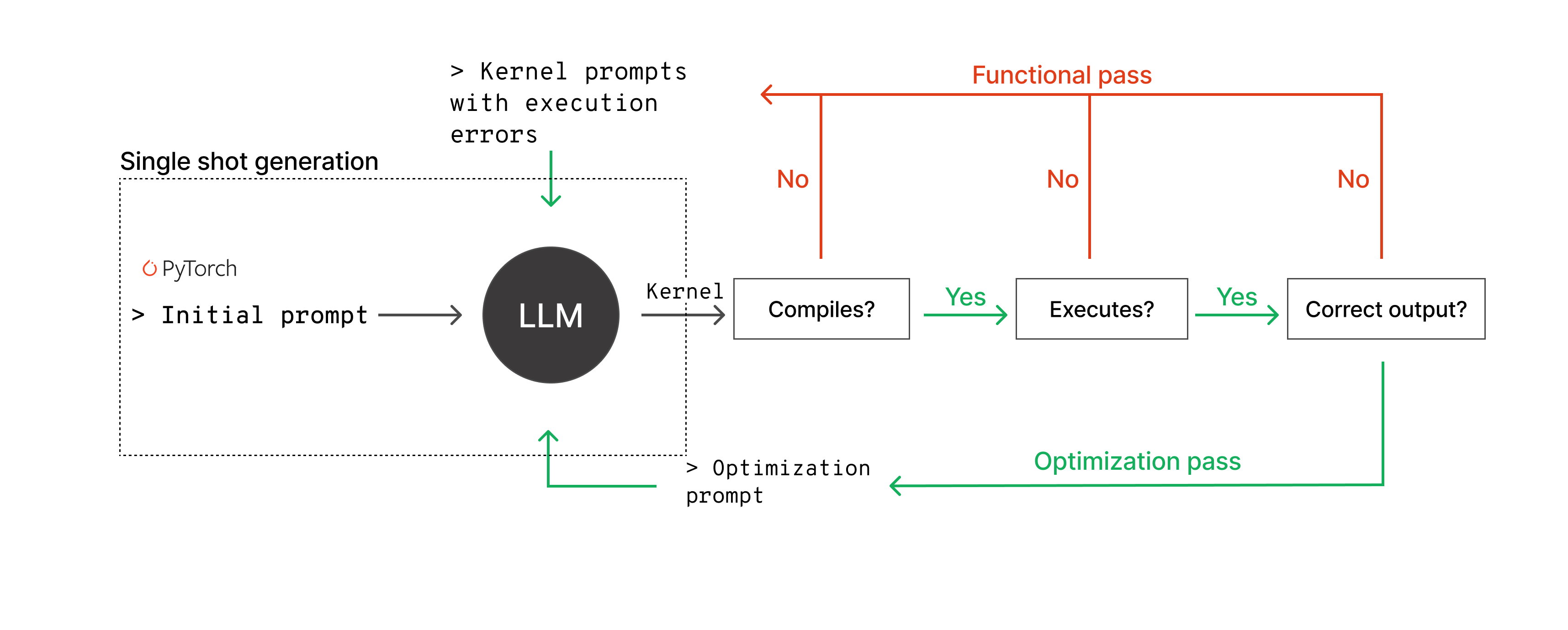}
    \vspace*{-15mm}
    \caption{Iterative program synthesis and optimization loop using LLMs. The workflow consists of two main phases: (1) a functional pass that iteratively refines synthesized programs until the code compiles, executes without errors, and produces correct output, and (2) an optimization pass that provides performance feedback to the LLM for iterative performance improvement. } 
    \label{fig:exec-flow}
\end{figure*}

\section{Introduction}

The rise in agentic AI applications is reshaping how inference workloads are deployed. A single pipeline now interleaves  LLM calls with reasoning, data retrieval, tool invocations, and multi-agent coordination, each with different compute, memory, and synchronization profiles. 
No single accelerator is the right fit for all of these stages at once. Production deployments hence need to assign each stage to its best-suited accelerator for optimal performance. 
This heterogeneity gives rise to a corresponding systems design challenge~\cite{gimlet}. 

Realizing this mapping across heterogeneous platforms requires optimized implementations of the same core operators across many different devices and programming models. Writing high-performance compute kernels requires mastering parallel programming languages such as CUDA~\cite{cuda}, 
Metal~\cite{apple_metal}, Triton~\cite{10.1145/3315508.3329973}, SYCL~\cite{intel_sycl}, or CuTe DSL~\cite{nvidia_cute_dsl}. 
Porting kernels across accelerators is rarely a syntactic translation, as each platform exposes different compute capabilities, memory hierarchies, bandwidth limits, and communication costs, and as a result, the best implementation often differs across platforms.



Modern compilers and inference runtimes have made substantial progress in automating performance optimization. Systems such as \texttt{torch.compile}~\cite{10.1145/3620665.3640366} and TensorRT-LLM~\cite{nvidia_tensorrt_llm} improve neural network execution through graph optimizations and 
automatic kernel fusion.
Nevertheless, building high-performance kernels requires combining clever algorithmic techniques with careful hardware utilization, as demonstrated by works such as FlashAttention~\cite{dao2022flashattention, dao2023flashattention2}, where integrating online softmax~\cite{milakov2018onlinenormalizercalculationsoftmax} with tiled attention computation and hardware-specific instructions enables superior performance. 
Together, these optimizations reduce memory traffic, improve arithmetic intensity, and avoid scheduling overheads that general-purpose compiler passes may not eliminate. End-to-end model inference complicates the problem further. A kernel that is faster in isolation may not improve full-model runtime if it prevents graph-level optimizations, introduces layout or precision conversions, or shifts bottlenecks elsewhere in the pipeline. Conversely, a kernel that appears modest in standalone benchmarks may unlock larger gains when composed with favorable surrounding operations. 
Kernel optimization must be evaluated both locally, at the level of individual kernels, and globally, in the context of full-model execution. 

Recent progress in Large Language Models (LLMs) has made code generation increasingly plausible. However, extending these capabilities to kernel generation remains a challenge. Low-level performance code is generally brittle: small deltas in memory layout or numerical precision can break correctness or eliminate performance gains. Furthermore, training data for LLMs is skewed toward CUDA, leaving emerging platforms underrepresented. 



In this work, we present KForge, a cross-platform framework for LLM-driven kernel generation and optimization, as described in Figure~\ref{fig:exec-flow}. 
KForge mirrors the real-world workflow of a kernel engineer through an iterative refinement loop. 
The system alternates between functional passes, which drive a candidate towards correctness, and optimization passes, which improve runtime once a correct implementation is reached. During functional passes, the generation agent proposes candidate kernels and uses compilation and correctness feedback to converge on a valid implementation. 
During optimization passes, the performance-analysis agent interprets profiling data, including both programmatic metrics and GUI-based profiler outputs, and produces concrete code-level recommendations. These recommendations target hardware utilization metrics such as memory bandwidth utilization, warp occupancy, and arithmetic intensity, and feed into the next round of synthesis.

KForge is architected to generate kernels across diverse hardware backends and programming models. We evaluate the same agentic architecture across various backends and programming models, 
enabling examination of when kernel generation agents can transfer algorithmic ideas across platforms, and when they need backend-specific guidance to exploit hardware-specific capabilities. We evaluate KForge through two case studies chosen to probe very different baseline availability: 
NVIDIA B200, where KForge competes against years of vendor-tuned reference kernels in TensorRT-LLM, and Intel Arc B580, where comparable hand-tuned references do not exist, and the system instead serves as a bring-up tool for emerging hardware. 
In summary, the paper makes the following \textbf{key contributions}:
\begin{itemize}
    \item We introduce KForge, 
   a multi-stage autonomous program synthesis
framework 
in which the generation and performance-analysis agents collaborate to produce correct and optimized kernels using compilation, correctness, and profiling feedback. 
    \item We describe a system architecture that supports four accelerator vendors (NVIDIA, AMD, Apple, Intel) and six programming models (CUDA, Triton, CuTe DSL, HIP, SYCL, Metal) through a uniform program synthesis interface. 
    \item We present case studies on two representative backends: end-to-end speedup over TensorRT-LLM on NVIDIA B200, and Triton kernel generation on Intel Arc B580 where comparable hand-tuned references are unavailable. 
\end{itemize}
\section{Related Work}

\textbf{LLM-driven kernel generation.} A growing body of work uses LLMs to generate and optimize GPU kernels, predominantly targeting the NVIDIA ecosystem where training data on kernel development is abundant. Sakana AI's CUDA Engineer~\cite{lange2025aicudaengineer} used evolutionary search for automated CUDA kernel discovery. Follow-up analysis found that the system exploited evaluation framework vulnerabilities, 
illustrating the difficulty of robust evaluation in this space. CUDA-LLM~\cite{chen2025cudallmllmswriteefficient} develops a Feature Search and Reinforcement (FSR) framework that combines compilation, correctness, and profiling feedback to refine CUDA kernels. KernelBlaster~\cite{kernelblaster} augments a GPU coding agent with a persistent CUDA knowledge base accumulated across tasks. Autocomp~\cite{autocomp} extends LLM-driven optimization beyond NVIDIA, targeting tensor accelerators more broadly through planned hardware optimizations and hardware feedback. However, most works tend to focus on per-kernel speedups rather than measuring impact on full-model execution.


\textbf{Kernel-level benchmarks.} Several benchmarks have been developed to evaluate LLM-generated kernels. KernelBench~\cite{ouyang2025kernelbenchllmswriteefficient} introduced a benchmark framework with 250 PyTorch workloads to evaluate LLMs' ability to generate efficient GPU kernels. The benchmark uses a $fast_p$ metric measuring both correctness and speedup over baseline implementations. 
In contrast, NVIDIA's SOL-ExecBench~\cite{sol-execbench} focuses on evaluating the theoretical lower limit 
using an analytical pipeline which incorporates the workload's FLOP count and byte count coupled with the target hardware's peak bandwidth/FLOP capabilities. While these benchmarks have advanced per-kernel evaluation, they leave open the question of how LLM-generated kernels affect end-to-end model performance once integrated into production runtimes such as TensorRT-LLM, where graph fusion, autotuning, and surrounding kernel selection interact with the kernel under study.

KForge generates kernels across four AI accelerators and six programming models through a uniform interface, and evaluates the impact of generated kernels on full-model execution against integrated production runtimes such as TensorRT-LLM.


\section{\modelname{}: Autonomous Program Synthesis}

\begin{figure*}[t]
    \centering
    \includegraphics[scale=0.45]{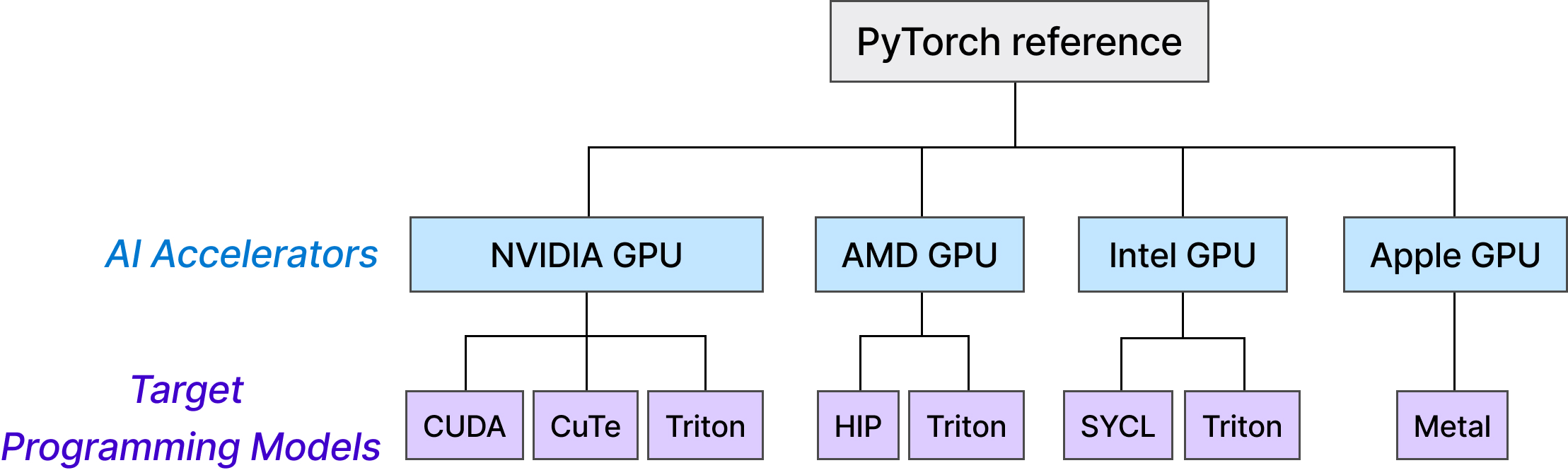}
    \vspace*{2mm}
    \caption{Given a PyTorch reference, KForge selects and lowers to an appropriate target programming model for each AI accelerator, supporting CUDA, CuTe, Triton, HIP, SYCL, and Metal across NVIDIA, AMD, Intel, and Apple hardware.} 
    \label{fig:backend-dsl}
\end{figure*}

\modelname{} is a multi-stage autonomous program synthesis framework, illustrated in Figure~\ref{fig:exec-flow}. It supports iterative refinement, single-shot program synthesis, as well as repetitive sampling, where the mode of operation is directed by prompt construction. It is designed as a cross-platform framework and currently supports a diverse set of hardware targets, spanning four vendors and six programming models with varying levels of abstraction, visualized in Figure~\ref{fig:backend-dsl}. For NVIDIA GPUs, we generate kernels in CUDA, Triton, and CuTe DSL, covering both low-level hand-tuned and higher-level abstractions. For AMD GPUs, we support HIP and Triton. For Intel Arc GPUs, we target SYCL and Triton. For Apple silicon, we generate Metal kernels. 
KForge is also model-agnostic; the underlying LLM is selected through a model registry, allowing users to plug in any model without changes to the synthesis pipeline. 

Within the scope of this work, we focus on the following three strategies that we employ for program synthesis. These strategies are complementary to one another, allowing one to build dynamic configurations based on available sources of supervision and computational resource budgets.

\begin{itemize}

    \item \textbf{Iterative refinement.} It allows the model to make error corrections from the previous run or optimize the performance of the correctly generated kernel, taking into account the program synthesized in the previous iteration. Specifically, for each iteration $i \in \{1, \ldots, N-1\}$ we add evaluation results from  iteration $i-1$ to the model's prompt, with a corresponding instruction to fix the error or improve program performance.

    \item \textbf{Cross-platform translation.} When a functional implementation already exists for one accelerator, it can be provided to the model as a reference, enabling cross-platform translation.
    For example, when generating Metal kernels, the prompt may include the corresponding CUDA implementation. This is particularly useful when the target backend lacks training data coverage. 
    
    \item \textbf{Profiling feedback.} Profiling data is crucial for pinpointing bottlenecks, providing comprehensive information on hardware resource usage for a specific computational workload. Timeline views assist in identifying scheduling gaps, while detailed statistics at the level of individual accelerator API calls highlight parts of the computational graph that do not fully utilize hardware resources.

\end{itemize}

\subsection{Program Synthesis Agent}
We follow a task definition similar to that in \cite{ouyang2025kernelbenchllmswriteefficient}. Specifically, we treat the LLM as a function $F: (p) \mapsto k$ that receives a text prompt $p \in \mathcal{T}$ as input and returns the generated code $k \in \mathcal{T}$. The generated code is expected to contain a kernel program, a kernel scheduling code, a JIT-library compilation code, and a PyTorch model \texttt{class NewModel(nn.Module)} with a \texttt{def forward(self, *inputs)} method that implements the module's forward pass. We use the Jinja2 template engine to parameterize the prompts. The default prompt $p$ contains a high-level task description, a one-shot example (a PyTorch implementation paired with its corresponding target-accelerator implementation), an input problem in PyTorch, and a task description in natural language.






\subsection{Performance Analysis Agent}

We introduce a specialized agent for performance analysis rather than handling generation and optimization in a single agent for two reasons.  First, profiling data is extensive but optimization signals are sparse; previous research~\cite{modarressi2025nolimalongcontextevaluationliteral} shows that LLM performance on relevant information retrieval drops to 50\% for 32K token inputs versus $<$1K tokens. Second, separating the two roles enables a modular architecture in which different models can be assigned to each agent based on their respective strengths. 


The Performance Analysis Agent processes profiling inputs, such as raw metrics from NVIDIA Nsight Systems or visual outputs from Xcode Instruments, and generates optimization recommendations for subsequent program synthesis iterations. This platform-agnostic approach handles arbitrary textual or visual profiling data across different hardware accelerators.

Formally, the agent is defined as $G: (o, k, \{v^{0},..., v^{n}\}) \mapsto r$, where $o \in \mathcal{T}$ is the text performance optimization prompt; $k \in \mathcal{T}$ is the synthesized program; $v^{i} \in \mathbb{R}^{H \times W \times C} \cup \mathcal{T}, i \in \{0,...,n\}$ represents profiling information as screenshots when $v^{i} \in \mathbb{R}^{H \times W \times C}$ or text-based profiler output when $v^{i} \in \mathcal{T}$; and $r \in \mathcal{T}$ is the performance recommendation. The agent is prompted to generate a single recommendation for maximum performance improvement.
The recommendation $r_t$ feeds into the next synthesis iteration, establishing a feedback loop: $F : (p, k_{t-1}, r_{t-1}) \mapsto k_t$.

\subsection{Program Verification}

The proposed execution flow defines a closed feedback loop, with information that is either helpful in recovering from failures or allows the model to optimize a functionally correct implementation to achieve speedup. After every generation-evaluation iteration, we save detailed logs for each workload. We focus on five possible execution states:
\begin{itemize}
    \item \textit{generation failure} --- typical reasons: network error, model output does not contain workload's code.
    \item \textit{compilation failure} --- the generated result contains workload's code, but fails to compile.
    \item \textit{runtime error} --- the workload's code compiles but fails at runtime, typically caused by segmentation faults or program abort.
    \item \textit{numerical or shape mismatch} --- call to \texttt{NewModel.forward} returns tensors, but they mismatch in tensor shapes or expected values, or both.
    \item \textit{correct} --- call to \texttt{NewModel.forward} returns tensors that match expected outputs both in shapes and numerically.
\end{itemize}

\subsection{Framework Features}
KForge includes several engineering features that make LLM-generated kernel optimization reliable, reproducible, and extensible. 
\begin{itemize}
    \item \textbf{Candidate guardrails.} KForge supports required and banned code patterns, allowing users to reject candidates that violate constraints. For example, default bans can prevent any undesired changes such as data type downcasting, fast-math flags, or changes to global precision settings.
    
    \item \textbf{Prompt editing.} KForge lets users override or append to the generation prompt globally or for specific iterations. This makes it possible to inject optimization hints, enforce experiment-specific instructions, or test alternative prompting strategies without modifying the core prompt templates.

    \item \textbf{Auxiliary code injection.} Users can provide auxiliary source files or directories to be included in all iterations or only selected iterations. This is useful for supplying helper kernels, prior implementations or any other reference code.

    \item \textbf{Structured artifacts for reproducibility.} Prompts, model responses, generated candidates, evaluation results, and profiling outputs are saved per model and per iteration, making experiments easier to inspect and reproduce.

    \item \textbf{Custom measurement hooks.} KForge supports user-defined action scripts that consume evaluation results and generated artifacts to compute task-specific measurements or scores, such as end-to-end runtime.
\end{itemize}

\section{Case Studies} 
We present two KForge case studies on backends with very different baseline reference availability: 
NVIDIA, where we compete against years of hand-tuned reference kernels, and Intel Arc, where comparable hand-tuned references are limited. For both cases, \modelname{} is powered by Claude Opus 4.6 in high-effort mode.
\label{sec:case_studies}

\subsection{GPT-OSS Optimization}
We target decode-path kernels in the modern MoE-based gpt-oss-20b architecture~\cite{openai2025gptoss120bgptoss20bmodel}, using TensorRT-LLM v1.3.0rc9 as our baseline. To identify optimization candidates, we profile an inference run with \texttt{nsys} and take the top 10 entries from the CUDA GPU Kernel Summary table. These fall into two categories: (1) proprietary kernels distributed only as cubins (e.g., attention and GEMM-with-activation fusions) and (2) kernels distributed with source code. We focus on category (2), since the available reference implementation provides a concrete starting point for optimization, and select three kernels: Fused Add + RMSNorm, MoE finalize, and Bias + RoPE + KV update. The chosen kernels together are components of the attention and MLP modules of the gpt-oss-20b MoE architecture.
 
In our experiments, we use the \texttt{trtllm-bench} script\footnote{\url{https://nvidia.github.io/TensorRT-LLM/commands/trtllm-bench.html}} in throughput mode with a workload of 512 requests, 1024-token prefill, 8192-token decode, and batch size 8, running 7 independent iterations of both the baseline and the \modelname{}-optimized model.
Initial profiling with \texttt{nsys} revealed high variance in time to first token (TTFT), time per output token (TPOT), and end-to-end latency across runs. We traced this to nondeterministic behavior in the autotuner, whose kernel selections varied from run to run. Disabling the autotuner and pinning the GPU clock to 1500 MHz produced reproducible latencies, and we use this configuration for both the baseline and \modelname{}-optimized model.

\subsubsection{Micro benchmarking}
To evaluate kernel performance across a range of workloads, we benchmark each kernel in isolation over a sweep of batch sizes. Results are reported in Table~\ref{tab:kernel_bench_sweep}. Our implementations are consistently faster than the baseline across nearly all configurations. The only exception is the Bias + RoPE + KV update kernel at batch size 8, where \modelname{} is marginally slower; at every larger batch size, the speedup grows with batch size. The MoE finalize kernel shows the most pronounced trend. While the speedup at small batch sizes is modest ($1.06$ to $1.10\times$), it increases to $1.43\times$ at batch size 128, indicating that our kernel scales better than the baseline as the workload grows.






\begin{table*}[h]
\centering
\small
\setlength{\tabcolsep}{5pt}
\caption{Per-kernel execution time ($\mu$s) for various batch sizes on NVIDIA B200.}
\label{tab:kernel_bench_sweep}
\begin{tabular}{lrrrrrrrr}
\toprule
\textbf{Kernel Name} & \textbf{Method} & \multicolumn{7}{c}{\textbf{Batch Size}} \\
\cmidrule(lr){3-9}
 & & \textbf{1} & \textbf{4} & \textbf{8} & \textbf{16} & \textbf{32} & \textbf{64} & \textbf{128} \\
\midrule
\multirow{3}{*}{Fused Add + RMSNorm}
 & Baseline        & 2.59 & 2.87 & 2.86 & 2.87 & 2.93 & 2.96 & 3.07 \\
 & \modelname{}    & \textbf{2.29} & \textbf{2.59} & \textbf{2.54} & \textbf{2.59} & \textbf{2.65} & \textbf{2.67} & \textbf{2.78} \\
 & Speedup         & 1.13$\times$ & 1.11$\times$ & 1.13$\times$ & 1.11$\times$ & 1.11$\times$ & 1.11$\times$ & 1.10$\times$ \\

\midrule
\multirow{3}{*}{MoE finalize}
 & Baseline        & 1.85 & 2.14 & 2.18 & 2.25 & 2.39 & 2.87 & 4.06 \\
 & \modelname{}    & \textbf{1.69} & \textbf{2.01} & \textbf{2.02} & \textbf{2.05} & \textbf{2.14} & \textbf{2.33} & \textbf{2.83} \\
 & Speedup         & 1.09$\times$ & 1.06$\times$ & 1.08$\times$ & 1.10$\times$ & 1.12$\times$ & 1.23$\times$ & 1.43$\times$ \\

\midrule
\multirow{3}{*}{Bias + RoPE + KV update}
 & Baseline        & 3.08 & 3.14 & \textbf{3.10} & 3.19 & 3.26 & 3.30 & 3.60 \\
 & \modelname{}    & \textbf{3.07} & \textbf{3.13} & 3.12 & \textbf{3.12} & \textbf{3.24} & \textbf{3.22} & \textbf{3.42} \\
 & Speedup         & 1.00$\times$ & 1.00$\times$ & 0.99$\times$ & 1.02$\times$ & 1.01$\times$ & 1.02$\times$ & 1.05$\times$ \\

\bottomrule
\end{tabular}
\end{table*}

\subsubsection{End-to-end performance evaluation}


On NVIDIA B200, integrating the optimized kernels yields a consistent end-to-end performance improvement. The results in Table~\ref{tab:e2e_results} are means over 7 independent iterations, with run-to-run coefficient of variation below $0.1\%$ for both metrics, more than an order of magnitude smaller than the reported gains. System output throughput improves by $2.12\%$, and total wall-clock time decreases by $2.07\%$. While modest in absolute terms, gains of this size are meaningful against a vendor-optimized runtime that already represents years of engineering effort, and they compound at deployment scale.

\begin{table}[h]
\centering
\small
\setlength{\tabcolsep}{6pt}
\caption{End-to-end performance on NVIDIA B200.}
\label{tab:e2e_results}
\begin{tabular}{lrrr}
\toprule
\textbf{Metric} & \textbf{Baseline} & \textbf{\modelname{}} & \textbf{$\Delta$} \\
\midrule
Throughput (tok/s)   & 2601.55 & \textbf{2656.61} & $+2.12\%$ \\
Wall-clock time (s)  & 1612.24 & \textbf{1578.82} & $-2.07\%$ \\
\bottomrule
\end{tabular}

\end{table}

\subsection{Triton Kernels on Intel Arc B580}
We evaluate \modelname{} on a non-NVIDIA platform by generating Triton kernels for the Intel Arc B580 (Battlemage). Our benchmark is the GEMM + tail-ops subset of KernelBench Level~2 (37 problems). Each problem runs five iterations of the generate-refine loop, and we compare the best generated kernel against the faster of two baselines: \texttt{torch.compile} and eager-mode PyTorch. \modelname{} achieves a $5.13\times$ geometric mean speedup across the 37 problems.

Table~\ref{tab:intel_b580_triton} shows four representative problems spanning the common tail-op patterns in this subset: pointwise activations with reductions (37, 62), normalization fused with elementwise ops (88, 62), and LogSumExp with activation chains (22). We manually inspected the kernels generated for these problems and find that they typically combine four techniques. \emph{Layer-wise fusion} collapses the post-GEMM sequence of operations into a single kernel that reads the matmul output once and writes the result once, eliminating the full-tensor round-trips that dominate baseline runtime. \emph{Mixed-precision execution} runs matmuls in FP16 on the B580's XMX units, with accumulation and post-processing in FP32. \emph{Single-pass reductions} replace two-pass GroupNorm and LogSumExp: small groups fit in registers so mean and variance accumulate alongside the data, while large reductions use the streaming online algorithm similar to FlashAttention's softmax. \emph{Partitioning choices}---sequential per-group loops, 2D group tiles, or per-row streaming---are tuned to group size and shape to maximize occupancy without cross-CTA (Cooperative Thread Array) reductions.
\begin{table}[h]
\centering
\small
\setlength{\tabcolsep}{3pt}
\caption{Execution time (ms) on Intel Arc B580.}
\label{tab:intel_b580_triton}
\begin{tabular}{lrrr}
\toprule
\textbf{Problem} & \textbf{\texttt{t.comp}} & \textbf{\modelname{}} & \textbf{Speedup} \\
\midrule
37\_Matmul\_Swish\_Sum\_GN     & 23.50 & \textbf{5.73} & $4.1\times$ \\
22\_Matmul\_Scale\_LSE\_Mish   & 9.86  & \textbf{1.51} & $6.5\times$ \\
88\_Gemm\_GN\_Swish\_Mul\_Swish & 10.00 & \textbf{1.65} & $6.1\times$ \\
62\_Matmul\_GN\_LReLU\_Sum     & 10.00 & \textbf{1.83} & $5.5\times$ \\
\bottomrule
\end{tabular}

\end{table}


\section{Limitations \& Future Work}
Several directions extend \modelname{} beyond the scope of this paper. The current implementation targets PyTorch as the source Deep Learning framework and verifies correctness through numerical tests with datatype-dependent tolerances. Broader framework coverage, such as JAX, and more rigorous verification approaches, such as differential testing across diverse input distributions or formal equivalence checking, would extend the system's applicability and strengthen confidence in LLM-generated kernels.

Beyond these, we highlight three research directions. First, the case studies in this paper rely on a source-level reference implementation to seed the generate-refine loop, but  many production kernels, such as fused attention, GEMM-with-activation cubins, and vendor BLAS, ship without source code. Synthesizing kernels from a behavioral specification alone, with correctness checked against the closed-source binary, is a harder regime that we plan to explore. Second, \modelname{} currently produces source-level kernels in  programming models such as CUDA C++ and Triton. Targeting a low-level virtual ISA such as PTX would unlock optimizations that source-level programming models cannot express, at the cost of portability and a more challenging synthesis problem. Third, isolated kernel speedups do not always translate to end-to-end gains, since a faster kernel can shift register and shared-memory pressure or perturb the autotuner's selections for neighboring kernels. 
We aim to extend \modelname{} to jointly optimize kernels within larger architectural components, such as attention or MoE MLP blocks, focusing on the entire block rather than optimizing each kernel independently. Finally, as agents become more capable of long-horizon reasoning, these directions can be further augmented by leveraging the recent advances in autonomous agentic approaches.


\section{Conclusion}

We presented \modelname{}, an agentic framework for cross-platform kernel generation and optimization. \modelname{} employs two collaborating LLM-based agents, a generation agent and a performance-analysis agent, to iteratively produce correct and performant kernels across NVIDIA, AMD, Intel, and Apple hardware through six different programming models. Two case studies demonstrate the framework under different baseline reference availability. On NVIDIA B200, \modelname{} produces end-to-end speedup over TensorRT-LLM, a strong baseline for gpt-oss-20b inference. 
On Intel Arc B580, \modelname{} generates Triton kernels for a backend that lacks a comparable hand-tuned reference. Vendor-optimized runtimes already represent years of engineering effort, and gains in the low single digits compound meaningfully at deployment scale. Agentic kernel synthesis is becoming a practical tool for both production-grade optimization and rapid bring-up on emerging hardware.


\bibliographystyle{IEEEtranS}
\bibliography{references}

\end{document}